\pdfoutput=1
\documentclass[conference]{IEEEtran}
\IEEEoverridecommandlockouts
\usepackage{cite}
\usepackage{amsmath,amssymb,amsfonts}
\usepackage{algorithmicx}
\usepackage{graphicx}
\usepackage{textcomp}
\usepackage{xcolor}
\usepackage{color}
\usepackage{ulem}
\def\BibTeX{{\rm B\kern-.05em{\sc i\kern-.025em b}\kern-.08em
    T\kern-.1667em\lower.7ex\hbox{E}\kern-.125emX}}
\begin{document}

\title{RRPN++: Guidance Towards More Accurate Scene Text Detection\\
}

\author{\IEEEauthorblockN{Jianqi Ma}
\IEEEauthorblockA{
mjq11302010044@gmail.com}

}

\maketitle

\begin{abstract}
RRPN is among the outstanding scene text detection approaches, but the manually-designed anchor and coarse proposal refinement make the performance still far from perfection. In this paper, we propose RRPN++ to exploit the potential of RRPN-based model by several improvements. Based on RRPN, we propose the Anchor-free Pyramid Proposal Networks (APPN) to generate first-stage proposals, which adopts the anchor-free design to reduce proposal number and accelerate the inference speed. In our second stage, both the detection branch and the recognition branch are incorporated to perform multi-task learning. In inference stage, the detection branch outputs the proposal refinement and the recognition branch predicts the transcript of the refined text region. Further, the recognition branch also helps rescore the proposals and eliminate the false positive proposals by the jointing filtering strategy. With these enhancements, we boost the detection results by $6\%$ of F-measure in ICDAR2015 compared to RRPN. Experiments conducted on other benchmarks also illustrate the superior performance and efficiency of our model.
\end{abstract}


\section{Introduction}

Text reading in scene image has been heatedly discussed in computer vision society for the past few years \cite{Ye2015Text,long2018scene,liu2019scene}. Great progress has been made in this area as numerous applications adopt text reading techniques such as document analysis, image retrieval and license template recognition system.

The general way of performing scene text reading can be consisted of two phases: detection and recognition. The detection part first retrieves the text region in the input image and extracted text regions are sent to the recognition part for the final text sequences. Therefore, the detection part acts as the first and significant role in scene text reading. However, current techniques still perform unsatisfactorily in challenging situation like {text scale variance}, low image resolution and complicated scene background \cite{Jung2004Text,Ye2015Text,liu2019scene}.

To tackle more complicated detection problem, traditional connected-component-based methods like \cite{CVPR2010SVT,2002BMVC_JMatas} try to find edges and pixels of the possible text instance and finally group them together to form a text-line region. They achieve superior performance in ICDAR 2011 \cite{Karatzas2011ICDAR} and ICDAR 2013 \cite{Karatzas2013ICDAR}.

\begin{figure}[t]
  \centering
  \includegraphics[width=\linewidth]{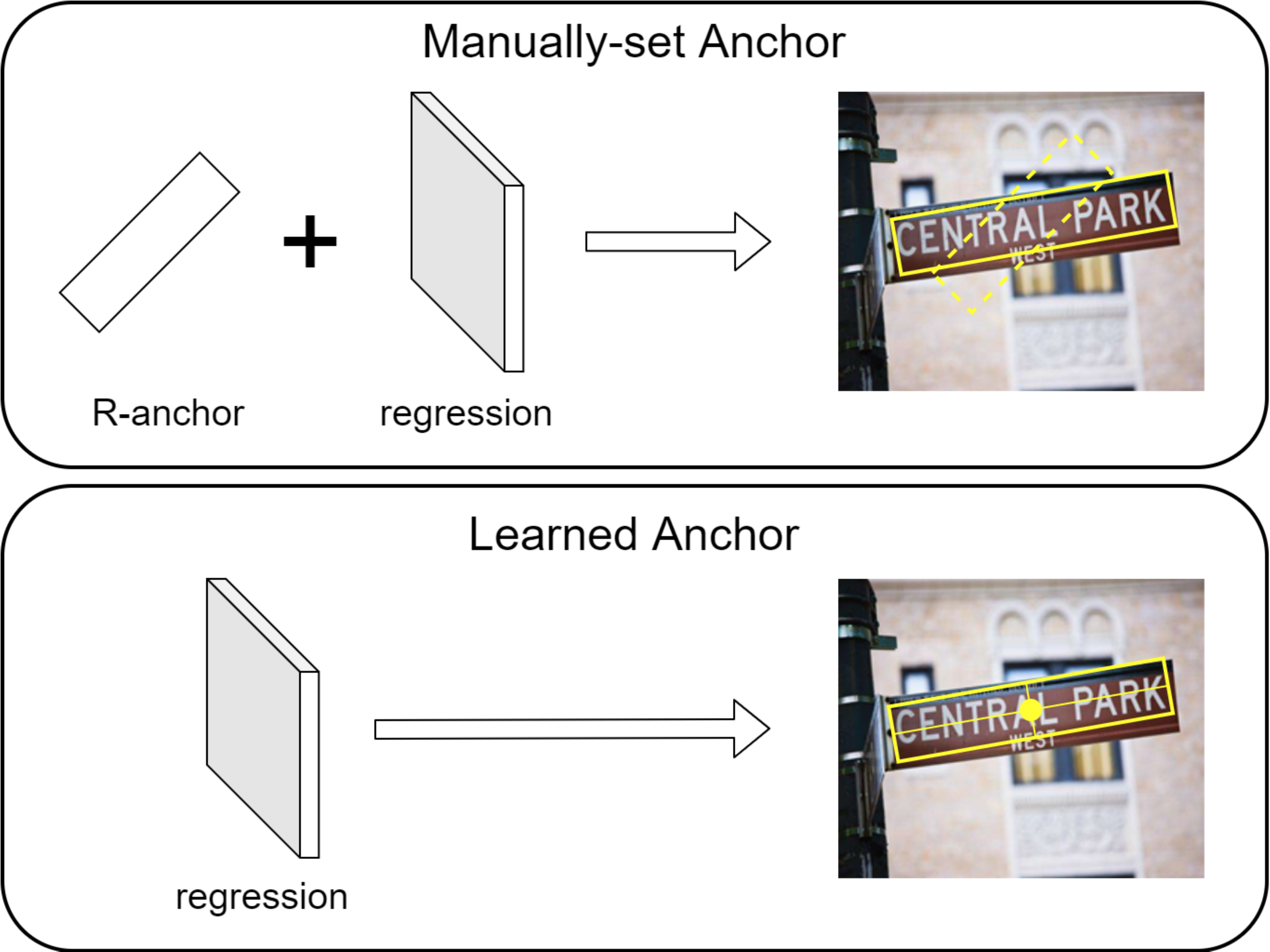}
  \caption{Comparison of manually-set anchor and learned anchor framework.}
  \label{fig:appn}
\end{figure}

As the neural network thrives since the time AlexNet wins the ImageNet challenge \cite{Krizhevsky:2012wla}, scene text detection has been pushed forward by CNN (Convolutional Neural Networks) methods \cite{tian2016detecting,zhang2016multi,Zhong2016DeepText} in axis-aligned scene text benchmarks like \cite{Karatzas2011ICDAR,Karatzas2013ICDAR}. Since the horizontal text detection approaches can easily adopt models from field of generic object detection, axis-aligned scene text is no longer too hard to detect. Therefore, research focus is shifted from frontal and horizontal text to a more general but challenging appearance such as perspective transformation, uneven light and length variation.

Anchor-based methods like \cite{liao2016textboxes,jiang2017r2cnn,liu2017deep,ma2018arbitrary} try to deal with the perspective distorted scene text with rotation proposal or quadrilateral boxes (see the top part of Figure \ref{fig:appn}). While segmentation-based approaches such as \cite{zhang2016multi,yao2016scene,PSENET,qiao2020text} try to solve the problem in pixel-level by learning the classification map and the boundary map. They employ polygon to represent a more accurate boundary of the text instance, so that the recognition model can have a better prediction after the polygon text region being rectified. Though segmentation-based methods can provide accurate boundary by generating polygons with numerous points, its inference speed is burdened by the post-process due to the fusion of massive factors such as irregular regions, boundary points and center line. Further, they may encounter performance drop when detecting large text  \cite{PSENET}. As to Anchor-based methods, the manually-designed anchor can cover most of the aspect ratios of the text instance, but the amount of anchors is so large that it could waste large amount of computation when dealing with the anchors (e.g., IoU computation and non-maximum suppression).

Meanwhile, the anchor-free concept\cite{huang2015densebox,tian2019fcos}  has been widely adopted in generic object detection which is also applied in scene text detection models. They tend to discard manually designed anchors and directly output the final detection instead of anchors with regression (shown in bottom part of Figure \ref{fig:appn}). Thus, the number of predicted boxes reduces from multiple anchors to one proposal per grid. What worths mentioning is that its label assignment becomes independent from IoU computation. Consequently, computation from IoU-based label assignment has been deprecated and the large Anchor-GT matrix is also no longer needed, which could occupy considerable amount of GPU memory and leads to the memory shortage.

Anchor-free methods in text detection like \cite{zhou2017east,liu2018fots,li2018pixel,deng2019stela} aim to predict the final detection by one regression. However, due to the uneven length of the text line presented, one-regression design always has weaker performance in text line detection. Variant length and scale further worsen the detection results. Thus a multi-scale adapted design \cite{lin2017feature} is somehow needed in the model. Further, training detection model with recognition transcripts \cite{liu2018fots,8578625inclinedE2E} also reports apparent performance improvement, in which the recognition branch can be considered a guidance to detection models. Hence, in RRPN++, we mainly discuss various guidance that can help improve the detection performance. Our contributions can be listed as follows:
\begin{itemize}
\item We propose an Anchor-free Proposal Pyramid Networks (APPN) to adapt text instance with various scales and release the detection learning from manually-designed anchors. The labels with different text instance of different sizes are assigned to different levels of APPN supervision so that the classification learning of multi-scale text can be balanced and efficient.
\item The factors that guide the text detection model towards higher performance are discussed in detail. They are: RRoI Align on extracting more accurate feature representation for the final detection prediction; Recognition branch on helping detection model obtain better performance with multi-task learning.
\item After comparing with current state-of-the-art approaches, we find RRPN++ outperforms most of the outstanding methods and finally shows its superiority in ICDAR2013, ICDAR2015, and COCO-Text.
\end{itemize}
\section{Related Work}

\subsection{Text Detection in the wild}

As the thrive of neural network in computer vision tasks, the most representative two-stage approach like Faster R-CNN\cite{Ren2016Faster} and one-stage model like SSD\cite{liu2016ssd} push the performance and speed forward respectively in generic object detection field. The axis-aligned bounding box is the output of the models which also related to the horizontal text in the natural scene. Hence Textboxes\cite{liao2016textboxes} and DeepText\cite{Zhong2016DeepText} are proposed based on the architecture to detect the horizontal text instance and achieve the state-of-the-art performance in ICDAR 2011\cite{Karatzas2011ICDAR} and ICDAR 2013\cite{Karatzas2013ICDAR} benchmarks.

However, horizontal text is not among the most common type of text appearances. Generally, the appearance of text instance suffers from perspective distortion and multi-orientation. Thus horizontal bounding box is unsatisfactory in dealing with text instance in these situations. Anchor-based methods RRPN\cite{ma2018arbitrary}, R2CNN\cite{jiang2017r2cnn} and DMPNet\cite{liu2017deep} propose rotation and quadrilateral anchors to adapt the text in multi-orientation in a tighter area. With the tighter detection representation, the recognition part can better predict the text sequence (reported in \cite{8334248Textboxes}). While anchor-free methods like EAST \cite{zhou2017east}, Pixel-Anchor \cite{li2018pixel}  and Learned-Anchor \cite{deng2019stela} directly learn bounding boxes from each grid point in feature map instead of difference between ground truth box and anchor.

Originally, the text illustrated in an image can be further various and irregular. Thus BPDN\cite{wang2019all} aims to output polygon to adapt the arbitrary-shaped text instance. With the points in polygon, text instance can be rectified into recognition-friendly shape by thin-plate-spline-based transformation.

\subsection{Guiding Proposals Towards Accurate Detection}

Guidance for detection and rescoring also plays an important role in detection tasks and reveals remarkable improvement in object detection and segmentation fields \cite{wang2019region,yang2018metaanchor,jiang2018acquisition,huang2019mask}.

\subsubsection{Anchor Guidance}

Anchor-based methods like \cite{ma2018arbitrary,jiang2017r2cnn,liu2017deep} manually design the anchor with fixed shape based on the training data distribution and human prior, i.e., we tend to set the aspect ratios towards prolate shape because the distribution of aspect ratio in the training set tends to be prolate. The model can converge better with properly set anchors. Meanwhile, the proposal of anchor-free idea enlarges the variance that the regression can adapt. Instead of predicting the bounding box difference, the regression head directly outputs the bounding box even if the text instance is of various length. Thus it can be regarded as a guidance of automatically adjustment of different aspect ratio. \cite{deng2019stela} generates the proposal by two branches, one for the centerness of the bounding boxes and the other for the width and height so that the model can accurately predict the region of text instance within fewer proposals. \cite{zhou2017east,li2018pixel} insist that each pixel in the feature map straightly output the rotation proposal with the distance to each ground truth edges or bounding points which also inspire the anchor-free construction of our APPN.

\subsubsection{Rescoring Detection}

The rescoring mechanism is first proposed in object detection and provide a novel thought in how to score the detection results. IoUNet \cite{jiang2018acquisition} applies one more branch in the second stage of detection model to directly predict the IoU value of each RoI. Moreover, the predicted IoU score also participates in NMS process. With these improvements, notable performance gain is observed by the conducted experiments. MaskScoring R-CNN\cite{huang2019mask} also adopts extra branch to predict the mask IoU which shows an extensive capability in boosting the performance of instance segmentation and detection by rescoring the predictions. While in fields of end-to-end text spotting, \cite{8237822horizontalE2E,lyu2018mask,liu2018fots} also design extra branch in the second stage of the model to predict the text sequence in the proposal. With extra supervision provided from recognition branch, the detection results are further improved by a clear gap. However, the scores in the recognition branch has rarely been applied to be a guidance of the detection results. Thus, besides the multi-task learning in detection gain, we also try to figure out whether the recognition can rescore the text instance in helping the text detection performance gain.

\section{Methodology}
In this section, we will first revise the RRPN\cite{ma2018arbitrary} model, and introduce our model architecture.

\subsection{Revising RRPN}\label{AA}

RRPN\cite{ma2018arbitrary} is a classic deep model for detecting arbitrary-oriented objects, especially in scene text detection. The model is first proposed to solve the perspective distortion of the scene text instance, and create rotated bounding box candidates to better fit the text instance. Based on Faster-RCNN\cite{Ren2016Faster}, RRPN imports the angle term into the anchor representation and learning. At the first stage, RRPN generates $54$ R-anchors ($6$ angles, $3$ scales and $3$ ratios) for each gird in the feature map. For the second stage, rotation proposals are filtered and taken as input to the R-CNN\cite{Girshick2015Fast} subnet. Rotated-ROI pooling \cite{ma2018arbitrary} is adopted to extract the corresponding Rotated-RoI feature from the $4$th stage of the backbone network. The cropped feature is sent into $2$ fully-connected layers, then the final layers output the regression prediction and classification scores of each proposal. Regression losses for both detection stages are smooth-L1 \cite{Ren2016Faster} loss and classification losses are cross-entropy loss.

\subsection{Anchor-free Pyramid Proposal Networks}

\begin{figure*}[t]
  \centering
  \includegraphics[width=\linewidth]{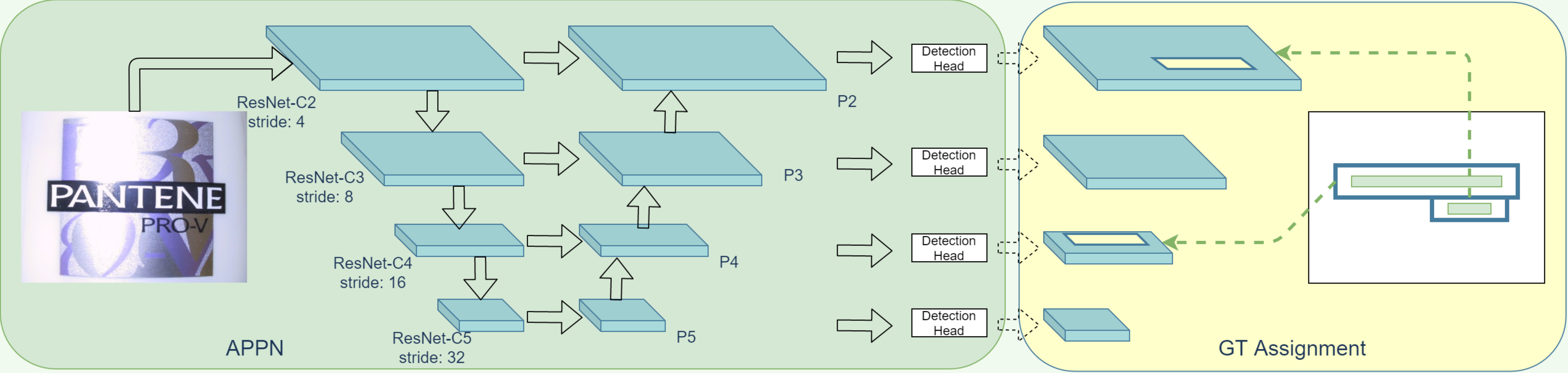}
  \caption{Pipeline of APPN and its ground truth assignment.}
  \label{fig:appnStructure}
\end{figure*}

Since anchor-free detection models become popular in recent years, approaches like \cite{zhou2017east,liu2018fots,li2018pixel,wang2019region} also achieve competitive results compared with anchor-based methods. Inspired by the construction of anchor-free framework, we change the representation term of the rotate boxes of $(x, y, w, h, \theta)$ to $(l, t, r, b, \theta)$, i.e., for gird point $(x, y)$ in a feature map, $l$, $t$, $r$, $b$ stand for distance (e.g. Euclidean distance) from point $(x, y)$ to left, top, right and bottom edges namely. While $\theta$ in both representations are the same, i.e., orientation of the box. This design is also introduced in EAST \cite{zhou2017east} and FOTS \cite{liu2018fots}.

Without predefined anchors to match objects, the network directly output the $5$ terms $(l, t, r, b, \theta)$ for each grid instead of the target shift used in RRPN. It largely reduces the number of the proposals to $\frac{1}{54}$ of the amount compared to RRPN on a feature map with the same size (i.e., we generate an $H \times W$ feature map with height $H$ and width $W$ while RRPN generates $54 \times H \times W$). Therefore, large amount of computing resource is saved when processing the proposals.

FPN \cite{lin2017feature} aims to balance the learning of bounding boxes with huge scale variance. It generates small-scale anchors to match small objects and large-scale for the large objects. The framework learns small objects in high resolution feature map and large objects in low resolution feature map. By grouping objects in similar scales in the same head, FPN better models objects of rapid scale variance. Thus we follow FPN to construct Anchor-free Pyramid Proposal Networks (APPN) on detecting multi-scale text instance. Here we pick 4 heads from different levels of the network stage. Namely, $P2$, $P3$, $P4$, $P5$ represent the feature map down-sampled from original input to scales of $\frac{1}{4}$, $\frac{1}{8}$, $\frac{1}{16}$ and $\frac{1}{32}$ respectively. Details can be viewed at Figure \ref{fig:appnStructure}. We choose a classic ResNet-50 \cite{he2016deep} to be the backbone of APPN. Then internal feature from FPN is set to 64 in channel number, after a Conv layer with kernel $3 \times 3$, output is branched into objectness and regression subnets. The objectness branch is a $1$-layer conv and outputs $1$-channel score map, the more the region is text-like the higher the score it will predict, while regression branch is also a $1$-layer conv to output a $5$-channel regression map which is normalized by sigmoid function. Here the sigmoid function will normalize the proposal terms into range $(0, 1)$. Thus we need to enlarge the range by a base size. Thus we will multiply the proposal terms with the base size (usually set to 640). It means that the proposals terms can range from $0$ to $640$. Each head outputs both the objectness scores and 5 terms of the bounding boxes.


\subsection{Rotated RoI Align}

RRPN modifies the RoI Pooling layer \cite{Girshick2015Fast} to adapt arbitrary-oriented proposals and extract the features of proposals from the global feature map, which is called RRoI Pooling. The rotated proposal is divided into $h \times w$ grids equally. RRoI Pooling first computes the corresponding proposal grid coordinates in the global feature map. After the boundaries of each grid are calculated. Max pooling operation is performed to determine the final value of the grid.

However, when the proposal is extremely large, it will cover large area of the global feature map and so does the proposal grid. Feature calculated near center at each divided grid using bilinear interpolation may not be strong enough to represent the textness in the grid. Further, RoI Align illustrates superior performance in Mask R-CNN \cite{he2017mask} due to its fine-grained grid sampling strategy. Besides dividing proposals into grids, RoI Align also samples at equal spatial intervals. The feature within the grid is calculated by a mean pooling of sampling points. Thus the larger amount of points it samples, the more representative the feature can be. We take the advantage of both rotation and RoI Align to extract more representative feature from rotation proposal. The operation has been implemented in $detectron2$ \cite{wu2019detectron2} project.

\subsection{Fast R-CNN branch}

 \textbf{Detection Head} After Rotated RoI Align layer, cropped feature batches are sent into the Fast R-CNN stage, here we build two subheads to perform different tasks. One subhead is adopted to predict the regression of the proposals from APPN. In the subnet, Rotated RoI features are flattened and sent into $2$ successive fully-connected (FC) layers with output size of $1,024$ and $2$ parallel FC layers to predict the final text score and box regression.

 \textbf{Recognition Head} Besides detection subnet, we also build a recognition branch to further predict the text sequence contained in the rotation proposal. Thus the Rotated RoI features are also taken as input of the recognition branch which may be extracted with different resolution from detection branch (e.g., we extract feature of $14 \times 14$ resolution in detection branch but of $8 \times 35$ for recognition branch).
We regard Conv-BN-ReLU as a single unit in the recognition branch, the connection of the subnet can be viewed at Table \ref{table:recognitionHead}. After $2$ Conv-BN-ReLUs, we downsample the feature in y dimension using a convolution layer with stride $(2, 1)$ to half the feature height. After each downsampling, the channel size is doubled. When the height of feature is downsampled to $1$, we adopt a bi-LSTM layer to encode the contextual semantics of the convolutional feature. Finally an FC layer takes the output from bi-LSTM to compute the probabilities of character sequence. With alphabet $S$, recognition head outputs probability vectors with size $|S|$.

\begin{table}[t]
\centering
\caption{Layer order and configurations of the recognition head. The $k$, $s$ and $p$ means kernel, stride, and padding set in convolution layer. The $out$ means output of the $Conv$ and $BN$ layer; $|S|$ represents the cardinal number of the total alphabet; If $k$, $s$ and $p$ are set in tuple $(h, w)$, $h$ means settings in $y$ axis while $w$ in $x$ axis}
\begin{tabular}{cc}
\hline
Layer Type & Parameter\\\hline
Input feature & C=64; W=35; H=8\\
Conv-BN-ReLU & $out$:64; $k$:(3, 3); $s$:1; $p$:1\\
Conv-BN-ReLU & $out$:64; $k$:(3, 3); $s$:1; $p$:1\\
Conv-BN-ReLU & $out$:128; $k$:(3, 3); $s$:(2, 1); $p$:1\\
Conv-BN-ReLU & $out$:128; $k$:(3, 3); $s$:1; $p$:1\\
Conv-BN-ReLU & $out$:128; $k$:(3, 3); $s$:1; $p$:1\\
Conv-BN-ReLU & $out$:256; $k$:(3, 3); $s$:(2, 1); $p$:1\\
Conv-BN-ReLU & $out$:256; $k$:(3, 3); $s$:1; $p$:1\\
Conv-BN-ReLU & $out$:256; $k$:(3, 3); $s$:1; $p$:1\\
Conv-BN-ReLU & $out$:256; $k$:(3, 3); $s$:(2, 1); $p$:1\\
bi-LSTM & hidden unit:256\\
FC & out:$|S|$\\
\hline
\end{tabular}
\label{table:recognitionHead}
\end{table}

\subsection{Ground Truth Generation and Learning}

\subsubsection{APPN}
In the first phase of RRPN, the positive samples are chosen by an IoU threshold of 0.7. By combining with negative samples, a balanced batch is grouped and makes it easier to converge in the objectness branch. While only positive anchors are learned in the regression branch. APPN tries to learn the regression without dense anchors. Thus using IoU-based strategy is so strict for choosing the positive samples that proposals generated by APPN can hardly match at the starting of training steps. Therefore, we directly fill a $1$-channel ground truth heatmap with manually assigned positive label. In the heatmap, grid filled with label $1$ is for positive samples which is inside the boundary of box and $0$ for negative samples outside. Shrinking is required for each box area to prevent overlapping for text instance closer to each other and we shrink the boundary to $0.7$ of its original. In order to better catch overall shape of the text instance, dice loss \cite{milletari2016v} is applied as follows:

\begin{equation}
\begin{array}{l}
    L_{cls}(y_{pred}, y_{gt}) = 1 - 2\frac{|y_{pred}||y_{gt}|}{|y_{pred}| + |y_{gt}|}
\end{array}
\end{equation}

$y_{pred}$ means the classification map that predicted by all detection heads of APPN, its calculation can be written as follows:

\begin{equation}
\begin{array}{l}
    y_{pred} = \mathop{\bigodot}\limits_{\{x_i \in X_L\}}{x_i} \\

\end{array}
\end{equation}

$X_L$ means all levels and output from all levels $L$, $\bigodot$ represents the concatenation operation. While the generation of $y_{gt}$ classification map can be written as follows:

\begin{equation}
\begin{array}{l}
    y_{gt} = \mathop{\bigodot}\limits_{\{r_k \in G_L, L_{i} \in L\}}{f(r_k, L_{i})}
\end{array}
\end{equation}

$G_L$ in the equation means the GT box sets that match specific levels of map. For the sake of convenience, GT boxes are grouped by the box area, i.e., $(0, 64^2]$, $(64^2, 128^2]$, $(128^2, 256^2]$ and $(256^2, +\infty)$ are namely assigned from P2 to P5. $f$ means the generation computation of the GT map.

As for the regression branch, we generate a regression map of 5 channel for each image whose channels stand for terms $R(l, t, r, b, \theta)$ of the box representation namely. By learning the $B(l, t, r, b)$ terms of the boxes, $IoU$ loss \cite{yu2016unitbox} is adopted to calculate the regression loss, while the loss of the angle can simply be computed using L1 loss. Therefore, total loss computation can be concluded by:

\begin{equation}
\begin{array}{l}
    L_{reg} = IoU(B_{gt}, B_{pred}) +
    \lambda_{\theta}|\theta_{gt} - \theta_{pred}|
\end{array}
\end{equation}

 $B_{pred}$ represents the regression maps of $(l, t, r, b)$ terms in all heads and $B_{gt}$ is the regression map generated by GT boxes in all levels. The angle in prediction and ground truth are $\theta_{pred}$ and $\theta_{gt}$ respectively. Hence, the learning of APPN phase consists of two parts: objectness and regression.

\begin{equation}
\begin{array}{l}
    L_{\text{APPN}} = \lambda_{\text{obj}}L_{\text{obj}} + \lambda_{\text{reg}}L_{\text{reg}}
\end{array}
\end{equation}

For the best performance, $\lambda_{obj}$, $\lambda_{reg}$ and $\lambda_{\theta}$ are set to $0.01$, $1$ and $20$ namely.

\subsubsection{Fast R-CNN} In Fast R-CNN stage, we follow the sampling strategies of RRPN. By using the IoU and angle difference, we filter the proposals from APPN. Here we manually define the positive sample to be: (1) rotated IoU with GT box is greater than 0.6 \textbf{and} (2) absolute angle difference with GT box is less than $\frac{\pi}{6}$; while the negative samples have the definition: (1) rotated IoU with GT box is less than 0.6 \textbf{or} (2) absolute angle difference with GT box is greater than $\frac{\pi}{6}$.

In detection head, both positive samples and and negative samples are needed for the learning of box classification. Generally, we group a batch of 256 RoI features, in which the positive samples have the amount of up to one quarter of the batch and negative samples take the rest. The classification loss is calculated using cross-entropy $CE$:

\begin{equation}
\begin{array}{l}
    L_{\text{Fcls}}(c, c^{*}) = \frac{1}{N}CE(c, c^{*})
\end{array}
\end{equation}

where $c$ and $c^{*}$ mean the predicted class probability and ground truth one-hot vector respectively, while $N$ refers to the number of a batch that contains positive and negative RoI samples.

The regression outputs the different of proposal and ground truth box which follows \cite{ma2018arbitrary}. Thus the box representation in grid coordinate $G(x_g, y_g)$ should first be changed from $R_{d}=(l, t, r, b, \theta)$ to $R_{r}=(x, y, w, h, \theta)$ for convenience of computation. By using the matrix multiplication, $R_{r}$ can be calculated as:

\begin{equation}
\begin{array}{l}
    (R_{r})^{T} = \mathbf{\Gamma} \cdot (\{R_{d};1\})^{T}
\end{array}
\end{equation}

where the semicolon in the equation means the concatenation operation and the transformation matrix $\mathbf{\Gamma}$ can be written by:

\begin{equation}
    \mathbf{\Gamma}=  \left[
\begin{array}{llllll}
\frac{-cos\theta}{2} & \frac{sin\theta}{2} & \frac{cos\theta}{2} & \frac{-sin\theta}{2} & 0 & x_g\\
\frac{-sin\theta}{2} & \frac{-cos\theta}{2} & \frac{sin\theta}{2} & \frac{cos\theta}{2} & 0 & y_g\\
1 & 0 & 1 & 0 & 0 & 0 \\
0 & 1 & 0 & 1 & 0 & 0 \\
0 & 0 & 0 & 0 & 1 & 0 \\
\end{array} \right]
\end{equation}

Hence, the generation of regression target $v^{*} = (v^{*}_x, v^{*}_y, v^{*}_h, v^{*}_w, v^{*}_\theta)$ follows relus of RRPN, which is calculated between predicted box $R_{rp}=(x, y, w, h, \theta)$ and ground truth box $R_{rgt}=(x^{*}, y^{*}, w^{*}, h^{*}, \theta^{*})$:


\begin{equation}
\begin{array}{l}
    v^*_{x}=\frac{x^* - x}{w}, v^*_{y} = \frac{y^*-y}{h} \\
    v^*_{h}=\log\frac{h^*}{h},v^*_{w} = \log\frac{w^*}{w}, v^*_{\theta}=\theta^* \ominus \theta
\end{array}
\end{equation}

Also, the smooth-L1 is used to calculate the regression loss $L_{\text{Freg}}$. Hence, the regression loss calculation for a batch in second stage can be written as:

\begin{equation}
   L_{\text{Freg}} = \frac{1}{N_{+}}\sum_{N_{+}}{\text{smooth}_{L_1}(v_i^{*}, v_i)}
\end{equation}

where $N_{+}$ means the number of the positive samples in the Fast R-CNN stage and $v_i$ represents the regression predicted by detection head.

In the recognition phase, we see the problem as a sequence learning situation. Naturally, CTC loss \cite{graves2006connectionist} is applied to model the learning of the possible sequence in a proposal. Note a feature-and-sequence pair as $\mathbf{P} = \{(F_k, L_k)|k \in N_{+}\}$, where $F_k$ is the feature extracted from $k$th proposal in a batch from APPN, $L_k$ is the corresponding ground truth sequence of the $k$th box. Thus we can adopt negative log-likelyhood to format the objective function of $\mathbf{P}$:

\begin{equation}
   L_{\text{Frec}} = -\frac{1}{N_{+}}\mathop{\sum}_{F_k, L_k \in \mathbf{P}}{\text{log} p(L_k|l_k)}
\end{equation}

where $l_k$ is the predicted sequence by recognition head from cropped feature $F_k$.

Therefore, the loss function for the Fast R-CNN stage can be formatted as:

\begin{equation}
   L_{\text{F}} = L_{\text{Fcls}} + L_{\text{Freg}} + L_{\text{Frec}}
\end{equation}

\subsubsection{Multi-Task Learning} Finally, combining all losses together, the full form of our multi-task loss is:
\begin{equation}
   L_{\text{F}} = L_{\text{APPN}} + L_{\text{F}}
\end{equation}

\subsection{Inference}

\subsubsection{Pipeline}

In the training stage, the RRoI features applied in the detection branch and the recognition branch are both extracted using the proposal from APPN. The two branches are parallel in the training process to maintain a fast training speed. While in the testing phase, the RoI feature will be sent to detection branch first. After being refined in the second regression, the proposals are employed in extracting RoI feature of recognition branch. Thus the recognition branch will predict the text sequence of proposals in latest position with the final refinement.

\subsubsection{Recognition Rescoring and Joint Filtering}

In traditional two-stage detector, the testing phase directly uses the detection scores in Fast R-CNN stage to process NMS and filter the bounding boxes. However, the scoring and detecting branches infers in parallel. The score predicted by the second branch is inaccurate for the refined proposals as the proposal status has been changed. The design has also been mentioned in \cite{cai2018cascade}. Thus in our testing stage, the recognition score $s_{r}$ is also used in filtering final detection, which can be calculated as follows:

\begin{equation}
   s_{r} = \frac{1}{|T|}\mathop{\sum}_{t \in T}{Softmax(r_{t})}
\end{equation}

where $r_{t}$ means the recognition output of a predicted rotation proposal at sequence timestamp $t$. After $Softmax$ process in all timestamps, we define the final recognition score as their average probability. Therefore, the final detection set can be filtered by both scores:

\begin{equation}
\begin{array}{l}
   D_{r} = \{d(s_{r}, s_{d})| s_{r} > t_{r}, d(s_{r}, s_{d}) \in D\} \\
   D_{d} = \{d(s_{r}, s_{d})| s_{d} > t_{d}, d(s_{r}, s_{d}) \in D\}
\end{array}
\end{equation}

For each detection $d(s_{r}, s_{d})$ from the original result set $D$, we have bounding box $d$ with scores $s_{r}$ from recognition branch and $s_{d}$ from detection branch. $t_{r}$ and $t_{d}$ are the thresholds for filtering recognition score and detection score respectively. The sets $D_{r}$ and $D_{d}$ are the candidate boxes that filtered with recognition score and detection score. Thus we can have our final detection $D^*$ by merging the sets $D_{r}$ and $D_{d}$.
\begin{equation}
\begin{array}{l}
   D^* = D_{r} \bigcup D_{d}
\end{array}
\end{equation}

\section{Experiments}

In this section, we will introduce our details in performing experiments of our models and illustrates how it can reach a superior detection performance.

\subsection{Datasets}

\textbf{SynthText} \cite{Gupta16} contains over 850,000 synthesized images using designated engine. The data quantity is close to the real-world image and its large amount help pretrain many models. The dataset provides annotations in character level and word level in the form of boxes in quadrilateral and transcription.

\textbf{ICDAR2013} \cite{Karatzas2013ICDAR} focuses on both horizontal scene text detection and recognition in natural world. The dataset can be divided into training set of 229 images and test set of 233 images. Annotation of each images contains text boxes in axis-aligned rectangle and its text sequence.

\textbf{ICDAR2015} \cite{karatzas2015icdar}, which is also called Incidental Scene Text Challenge, mainly concentrates on detecting and recognizing text instances in arbitrary-orientation. The dataset contains a training set of 1,000 images and a test set of 500 images. All of the images are took incidentally in the streets and shopping malls in real world. Its annotation provides boxes in quadrilateral and transcription of each text instance.

\textbf{ICDAR17-MLT} \cite{nayef2017icdar2017} is a challenge aiming at multi-lingual scene text detection and recognition in real world. It has a training set of 7,200 images and a validation set of 1,800 images and a test set of 9,000 images. It is a dataset with multi-lingual and multi-oriented situation. Moreover, the various text length distribution in different languages makes it more challenging in detection task. The text regions are annotated in form of quadrilateral and the transcriptions are also given for recognition.

\textbf{COCO-TEXT} \cite{veit2016coco} is the largest text detection dataset in the real-world, which originates from MS-COCO \cite{lin2014microsoft}. All text instances in 63,686 images of MS-COCO are annotated. The training set has 43,686 images, while 10,000 of the rest 20,000 images are chosen to be test set and last 10,000 images to be the validation set. The dataset provides word boxes formed in quadrilateral in its latest version and the corresponding word content.

\subsection{Implementation Details}

\subsubsection{Training Hyperparameters}

We apply SGD optimizer to conduct our training process. In pretraining process, the learning rate is set to $2 \times 10^{-3}$ remains the same in the whole training process. While in the training process, we first set the learning rate to $2 \times 10^{-3}$ for the first 10 epoches, then reduce to $2 \times 10^{-4}$ for the next 10 epoches and $2 \times 10^{-5}$ for the final 5 epoches. The momentum is set to 0.9 and weight decay is 0.0005. Warming up strategy is used for the first 500 steps from one third of the learning rate, and grows in a linear manner. All experiments are conducted in a single GTX $1080$Ti GPU of CUDA $9.1$ and Pytorch $1.0$ version.

\subsubsection{Data Augmentation}

CutMix \cite{9008296cutmix} is originally applied in image classification tasks, it randomly cuts off a piece in one image, and fills the cut region with image patch from other images. While labels of the image are combined together using weighted sum.

Inspired by CutMix, we randomly crop image patches from different datasets and piece the patches together to form a larger size of input. The random crop ratio is chosen from interval of $[0.3, 0.7]$ proportion of an original image. Annotations of different patches are also combined together. While before sending the images into the model, we rescale the shorter side of image in one of $(640, 768, 896)$ while aspect ratio is kept. In order to balance the training speed and performance, Each input has 2 pieces of patches.

\subsection{Ablation Study}

To conduct our baseline experiments, we first process the pretraining step and train our model in SynthText dataset for nearly 5 epoches. Then in the training phase, the whole training set is formed by training/validation set from ICDAR2013, ICDAR2015, and ICDAR2017-MLT, in which some pictures without English transcription are removed. After 25 epoches of training, the baseline model continues to finetune in ICDAR2015 training set for another 20 epoches. Then we can have the baseline results in ICDAR2015 shown in Table \ref{table:ablationAlign}, and our following ablation studies are also conducted and validated in ICDAR2015.

\begin{table}[t]
\centering
\caption{Ablation on efficiency of feature pooling operation. We re-implemented RRPN models as $\text{RRPN}_{R}$.}
\begin{tabular}{c|c|c}
\hline
Approach & Pooling Operation & F-measure\\\hline
RRPN \cite{ma2018arbitrary} & RRoI Pooling &80.2\%\\
$\text{RRPN}_{R}$ & RRoI Pooling  &83.6\%\\
$\text{RRPN}_{R}$ & RRoI Align  &84.8\%\\
Baseline & RRoI Align &86.4\%\\
\end{tabular}
\label{table:ablationAlign}
\end{table}

\begin{table}[t]
\centering
\caption{Exploit the efficiency of data augmentation. Results are all validated on ICDAR2015 dataset}
\begin{tabular}{c|c|c}
\hline
Recognition Head & Score Filtering & F-measure\\\hline
$\times$ & $\times$ & 86.4\%\\
$\surd$ & $\times$ & 89.1\%\\
$\surd$ & $\surd$ & 89.5\%\\
\end{tabular}
\label{table:ablationFilter}
\end{table}

\subsubsection{RRoI Pooling vs. RRoI Align}

Rotated RoI Pooling (RRoI Pooling) \cite{ma2018arbitrary} aims to extract feature along orientation of rotation proposals. In order to solve the problems of feature inaccuracy caused by max pooling. Bilinear interpolation is adopted to alleviate the problems. However, feature extracted in center position of a large grid can hardly represent the textness in grid. In this case, dense sampling seems to be necessary. Here we conduct experiments to show the effectiveness of dense sampling in Table \ref{table:ablationAlign}. $\text{RRPN}_R$ means that we re-implement RRPN in pytorch version and use ResNet50 to be its backbone. After trained sufficiently, we finally have $83.6\%$ in F-measure. While we change the RRoI Pooling layer to RRoI Align layer, the experiment shows apparent performance gain by $84.8\%$. Further, our baseline is defined to be without recognition branch which achieves F-measure of $86.4\%$. Apparently, our APPN design boosts the detection performance by a clear gap compared to RRPN's anchor-based design.

\subsubsection{Joint Score Filtering}

 Recognition scores can be another important evidence for filtering the predicted detection set. Experiments in Table \ref{table:ablationFilter} also shows the improvement of the performance with the impact of Joint Filtering. Benefit from recognition learning, our detection obtains a $2.7\%$ performance gain and up to $89.1\%$. The final result further boosts to $89.5\%$ by applying joint filtering.

\begin{table}[t]
\centering
\caption{Results for ICDAR2013. 'DetEval' means the results follow the DetEval protocols in detection, while '$\text{WS}_S$' and '$\text{E2E}_S$' represent the Word-Spotting and End-to-End protocols in recognition. 'MS' means multi-scale testing and 'E2E' means end-to-end training with recognition. Upper part of the table shows the results of regression-based methods, while lower part shows segmentation-based results. Methods with '*' means recognition is not performed in end-to-end manner.}
\begin{tabular}{|c|c||c|c|c|c|}
\hline
\small{Approach} & Year & \multicolumn{1}{|c|}{F-measure} & \multicolumn{1}{|c|}{$\text{WS}_S$} & $\text{E2E}_S$ & Speed \\\hline\hline
\scriptsize{Textboxes++}* \cite{8334248Textboxes} & 2018 & 89.4\% & 96\% & 93\% & 11.6fps\\\hline
FOTS \cite{liu2018fots} & 2018 & 88.3\% & 92.7\% & 88.8\% & 7.5fps\\\hline
He et al. \cite{8578625inclinedE2E}& 2018 & 90\% & 93\% & 91\% &-\\\hline
BPDN E2E \cite{wang2019all} & 2019 & 90.1\% & - & 92.2\% & 4.6fps\\\hline
STELA \cite{deng2019stela} & 2019 & 91.5\% & - & - & 10.5fps\\\hline
Ours E2E & - & \textbf{92.0\%} & 92.9\% & 87.8\% & 13.3fps\\\hline\hline
\scriptsize{Mask TextSpotter} \cite{8812908masktextspotter} & 2018 & 91.7\% & 92.5\% & 92.2\% & 4.6fps\\\hline
PTMD \cite{liu2019pyramid} & 2019 & 93.6\%  & - & - & -\\\hline
TP E2E \cite{qiao2020text} & 2020 & 91.7\% & 94.9\% & 91.4\% & 8.8fps\\\hline
\end{tabular}
\label{table:ic13Results}
\end{table}

\subsection{Results on Benchmarks}

\subsubsection{ICDAR2013}
 We follow the pretrain and training step of the baseline experiments, then fine-tune the model in IC13 dataset for another 20 epoches. In inference phase of the model, we resize shorter size of the input image to $640px$ and keep ratio the same as its original. By adopting all strategies introduced in the last section, performance of our end-to-end model is shown in Table \ref{table:ic13Results}. Among all the regression-based methods, our model achieves $92.0\%$ f-measure and outperform all of its counterparts with the fastest inference speed of $13.3fps$. While compared with segmentation-based method PTMD\cite{liu2019pyramid}, our proposed model falls behind by $1.6\%$ but still leads in speed.

\subsubsection{ICDAR2015}
In the training stage, we keep all training steps the same with baseline experiments. While in the inference stage, we resize the shorter size of the image to $1440px$ and ratio of the images are kept. Our model also achieves a results of $89.5\%$ f-measure in a speed of $4.8$fps. which is over $0.9\%$ leads the second of regression-based method.

\subsubsection{COCO-Text}
In performance comparison of COCO-Text, besides the training sets mentioned in the baseline experiment, training set and validation set are also included. After the pretraining and training step, we input the image the same size as in IC15 comparison. We also reach state-of-the-art performance compared to all methods that listed in Table \ref{table:cocotextResults} and achieve a $1.6\%$ lead in f-measure.

\subsection{Discussion}
\begin{figure*}[t]
  \centering
  \includegraphics[width=\linewidth]{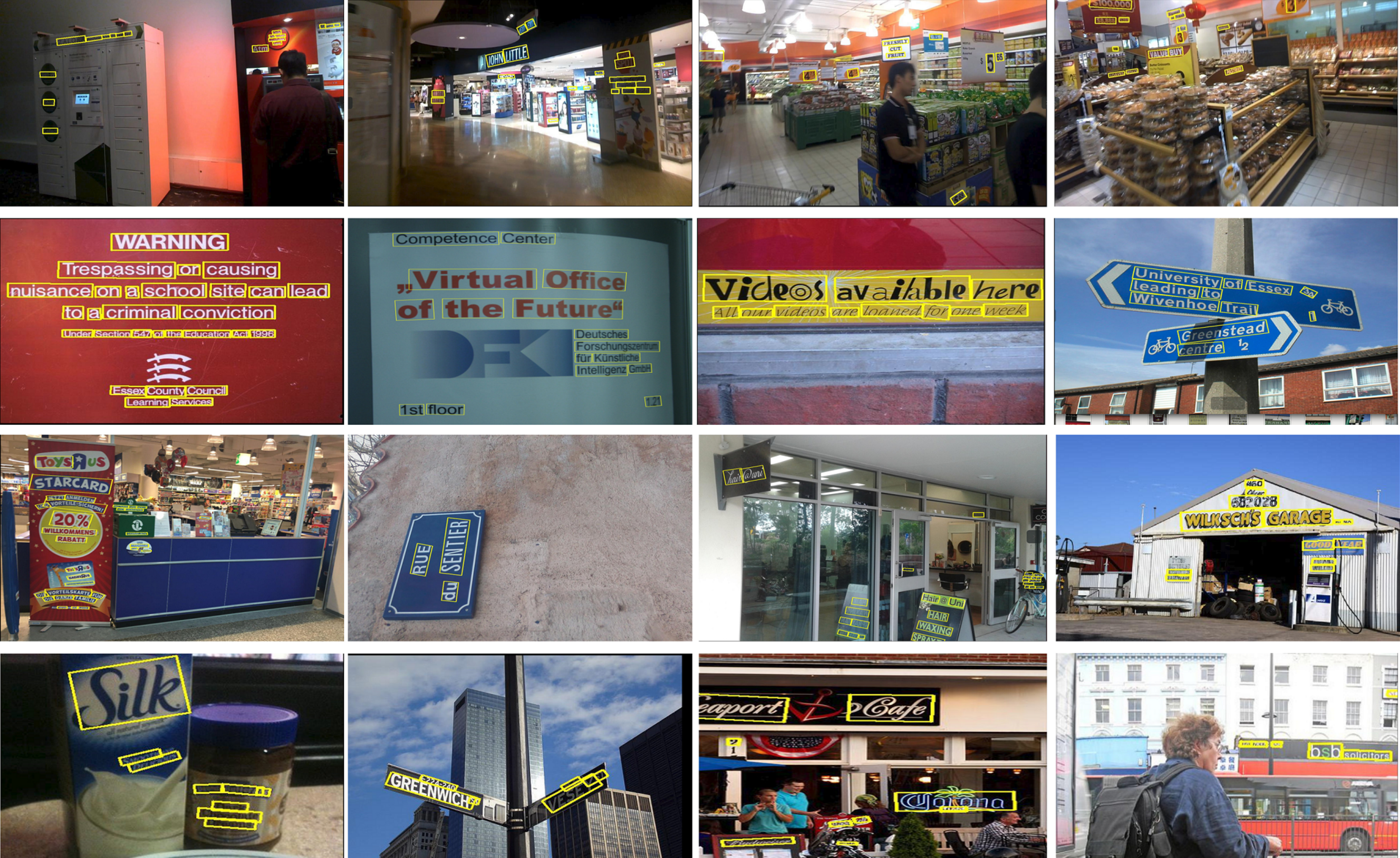}
  \caption{Visualization of detection performed by RRPN++. The visualized samples from top row to bottom are namely from ICDAR2015, ICDAR2013, ICDAR2017-MLT and COCO-Text.}
  \label{fig:visualization}
\end{figure*}

The detection results can be viewed at Figure \ref{fig:visualization}, which illustrates the robustness in handling small and multi-oriented scene text in several benchmarks. Since we design our recognition branch by just borrowing the architecture from CRNN \cite{shi2016end}, the recognition branch performs inferior among the methods. The assistance of boosting detection performance is apparent.

\section{Conclusion}

In this paper, we propose RRPN++ model to adopts anchor-free design to directly guide the learning of proposals in grid of feature map and feature pyramid structure to adapt scale variety of text instance. Further, detection performance is boosted with recognition supervision and joint filtering by detection score and recognition score. Experiments conducted in several benchmark also shows the superiority of proposed model.

\begin{table}[t]
\centering
\caption{Results for ICDAR2015. 'MS' means multi-scale testing and 'E2E' means end-to-end training with recognition.Upper part of the table shows the results of regression-based methods, while lower part shows segmentation-based results.}
\begin{tabular}{|c|c||c|c|c|c|}
\hline
\small{Approach} & Year & F-measure & $\text{WS}_S$ & $\text{E2E}_S$ & Speed \\\hline\hline
$\text{EAST}^{*}$\cite{zhou2017east} & 2017 & 80.7\% & - & - & -\\\hline
\scriptsize{Textboxes++} \cite{8334248Textboxes} & 2018 & 81.7\% & 76.5\% & 73.4\% & 11.6fps\\\hline
PixelAnchor \cite{li2018pixel}& 2018 & 87.7\% & - & - & -\\\hline
FOTS E2E \cite{liu2018fots} & 2018 & 88.0\% & 84.7\% & 81.1\% & 7.5fps\\\hline
He et al. \cite{8578625inclinedE2E}& 2018 & 87\% & 85\% & 82\% & -\\\hline
GNN \cite{9010643GNN} & 2019 & 88.5\% & - & - & 2.1fps\\\hline
BPDN E2E \cite{wang2019all}& 2019 & 88.6\% & - & 79.7\% & -\\\hline
Ours E2E& - & \textbf{89.5\%} & 83.6\% & 79.0\% & 3.5fps\\\hline\hline
\scriptsize{Mask TextSpotter} \cite{8812908masktextspotter} & 2018 & 86\% & 79.3\% & 79.3\% & 4.8fps\\\hline
Liu et al. \cite{liu2019omnidirectional}& 2019 & 86.5\% & & & -\\\hline
PSENet \cite{8953886PSENET}& 2019 & 87.2\% & - & - & 2.3fps\\\hline
Qin et al. E2E \cite{qin2019towards}& 2019 & 87.5\% & - & 83.4\% & -\\\hline
PTMD \cite{liu2019pyramid} & 2019 & 89.3\% & - & - & -\\\hline
TP E2E \cite{qiao2020text}& 2020 & 87.1\% & 84.1\% & 80.5\% & 8.8fps\\\hline
\end{tabular}
\label{table:ic15Results}
\end{table}

\begin{table}[t]
\centering
\caption{Results for COCO-TEXT. 'MS' means multi-scale testing and 'E2E' means end-to-end training with recognition.}
\begin{tabular}{|c|c|c|c|}
\hline
\small{Approach} & Year & F-measure & Speed \\\hline\hline
$\text{EAST}$\cite{zhou2017east} & 2017 & 39.5\% & 6.5fps\\\hline
Textboxes++ \cite{8334248Textboxes} & 2018 & 55.9\% & 11.6fps\\\hline
BPDN E2E \cite{wang2019all}& 2019 & 63.0\% & -\\\hline
Ours E2E& - & \textbf{64.6\%} & 5.2fps\\\hline
\end{tabular}
\label{table:cocotextResults}
\end{table}

\bibliographystyle{IEEEtran}
\bibliography{total}

\end{document}